\documentclass[runningheads]{llncs}

\usepackage{pifont} 
\usepackage{graphicx}
\usepackage{amsmath,amssymb}
\usepackage{booktabs}
\usepackage{multirow}
\usepackage{xcolor}
\usepackage{subcaption}
\usepackage{tikz}
\usepackage{pgfplots}
\usepackage[colorlinks=true, citecolor=blue!80!black, linkcolor=black, urlcolor=magenta]{hyperref}

\begin{document}

\title{TRIUNE-Net: Harmonizing Scale, Shape, and Efficiency in Pancreatic Tumor Segmentation}

\titlerunning{TRIUNE-Net}


\author{
Amir Hossein Saleknia\inst{1} \and
Alireza Kheyrkhah\inst{1} \and
Sanaz Karimijafarbigloo\inst{2} \and
Reza Azad\inst{2} \and
Sina Houshmand\inst{3} \and
Ulas Bagci\inst{4} \and
Dorit Merhof\inst{2} \and
Alaa Sulaiman \inst{5}
}

\authorrunning{A. Kheyrkhah et al.}

\institute{
\textsuperscript{1}Independent Researcher,
\textsuperscript{2}University of Regensburg, Germany,
\textsuperscript{2}University of California, San Francisco, USA,
\textsuperscript{4}Northwestern University, USA,
\textsuperscript{5}Abdora INC, USA\\
\email{alaa@abdora.ai}
}

\maketitle

\begin{abstract}

Pancreatic tumor segmentation in 3D CT volumes is challenged by extreme scale 
variability across both the pancreas and tumor, and highly irregular tumor 
morphology. While recent advances have pushed segmentation performance, existing 
methods do not explicitly address these challenges and come at the cost of 
excessive computational complexity, limiting their practicality in 
resource-constrained clinical environments. We propose TRIUNE-Net, a lightweight 
unified architecture that harmonizes scale, 
shape, and efficiency through three synergistic innovations. A multi-scale 
context aggregation module with stage-adaptive dilated convolutions enables 
the model to reason across the broad range of anatomical scales present in 
both organs. A serial linear-deformable attention mechanism combines large 
effective receptive fields with shape-adaptive deformable convolutions to 
capture irregular, non-convex tumor morphologies. Finally, an information-preserving 
downsampling module replaces conventional max pooling entirely, retaining all 
spatial information while adding negligible parameters, preventing small tumors 
from being discarded before they can be recognized. On both the MSD Pancreas and NVD Pancreas datasets, TRIUNE-Net achieves state-of-the-art results with only 5.86\,M parameters and no external pre-training, outperforming all baselines across all key tumor metrics. Specifically, it surpasses the next-best model by 0.45\% in tumor Dice, 
6.0 points in F1 score, 6.6 points in sensitivity, and 3.4 points in precision,
simultaneously reflecting its ability to suppress both missed tumors and 
false alarms in clinically realistic conditions. Our code is available at: \href{https://github.com/abdora-ai/TRIUNE-Net}{https://github.com/abdora-ai/TRIUNE-Net}


\end{abstract}

\section{Introduction and Related Work}

Pancreatic cancer is among the deadliest malignancies, with a five-year survival of only about 13\%, the lowest of any major cancer, ranking third in US cancer deaths~\cite{siegel2024cancer}. Yet automated pancreatic tumor segmentation remains notoriously difficult~\cite{antonelli2022medical}, a challenge stemming from three factors rarely addressed together: extreme scale disparity (the pancreas ranges 20--201 cm$^3$, tumors 0.4--732 cm$^3$, with early lesions under 0.01\% of the volume), highly irregular and non-convex morphology, and poor visibility of small tumors against surrounding tissue. These challenges are particularly acute in the opportunistic screening setting, where pancreatic tumors are often visible yet routinely missed in the tens of millions of abdominal CT scans acquired annually for unrelated indications~\cite{xia2022felix,cao2023large}. Seizing this opportunity requires models light enough to run on every study.

Research on 3D medical image segmentation has been driven mainly by accuracy, with most architectures built to generalize across organs and tasks. The dominant line is hybrid CNN-Transformer design: UNETR~\cite{hatamizadeh2022unetr} and Swin UNETR~\cite{hatamizadeh2021swin} couple transformer encoders with convolutional decoders to capture global context, while TransUNet~\cite{chen2021transunet} and UNETR++~\cite{shaker2024unetr++} streamline the attention interface; deformable convolutions and large-kernel attention~\cite{dai2017deformable,guo2023visual,azad2024beyond} add shape adaptivity, as in D-LKA Net~\cite{azad2024beyond}. These advances are real, but arrive at tens to hundreds of millions of parameters and frequently depend on large-scale pre-training. Lightweight designs prove efficiency is feasible: LHU-Net~\cite{sadegheih2024lhu} rivals far larger networks at a fraction of the cost, and MedNeXt~\cite{roy2023mednext} shows convolutional networks can remain competitive with substantially fewer parameters. Despite these advances, no existing lightweight architecture jointly addresses the three specific challenges of pancreatic tumor segmentation: extreme scale disparity, irregular morphology, and poor visibility of small tumors. Most methods treat these as incidental rather than primary constraints, resulting in computationally heavy or general-purpose solutions.

We present \textbf{TRIUNE-Net}, a lightweight U-shaped network with only \textbf{5.86\,M} parameters, designed to address all three challenges jointly and efficiently. Its core building block, the \textbf{Contextual Block}, comprises three complementary components: \ding{182}~a \textbf{Multi-Scale Gated Aggregator (MSGA)} using stage-adaptive dilated convolutions to model both global organ context and local lesions; \ding{183}~a \textbf{serial linear-deformable attention (D-LKA)} mechanism combining large receptive fields with deformable convolutions to adapt to irregular shapes; and \ding{184}~an \textbf{information-preserving downsampling (IPD)} module retaining all spatial information via space-to-depth rearrangement. On MSD and NVD Pancreas datasets, TRIUNE-Net delivers superior tumor segmentation without external pre-training, outperforming heavier models across all key metrics and proving that efficiency and accuracy aren't mutually exclusive.

\section{Method}

\subsection{Overview}

A lasting legacy of U-Net~\cite{ronneberger2015u} is to repeat the same block at
every resolution and mirror it symmetrically across the encoder and decoder. In 3D
this convenience is expensive and poorly matched to the data: the two
highest-resolution stages dominate the FLOP budget, because cost grows with spatial
extent, yet the features they produce are shallow textures that, in abdominal CT,
are mostly background and noise. The reasoning that actually separates a tumor from
surrounding parenchyma emerges only deeper, at coarser resolutions. We therefore
reject uniform design and \emph{allocate computation asymmetrically}: the
high-resolution stages act as a cheap transport layer that carries spatial detail
at minimal cost, while every parameter we spend on perception is concentrated in
the three coarse \emph{semantic} stages, where it does the most work.

\begin{figure}[t]
  \centering
  \includegraphics[width=\linewidth]{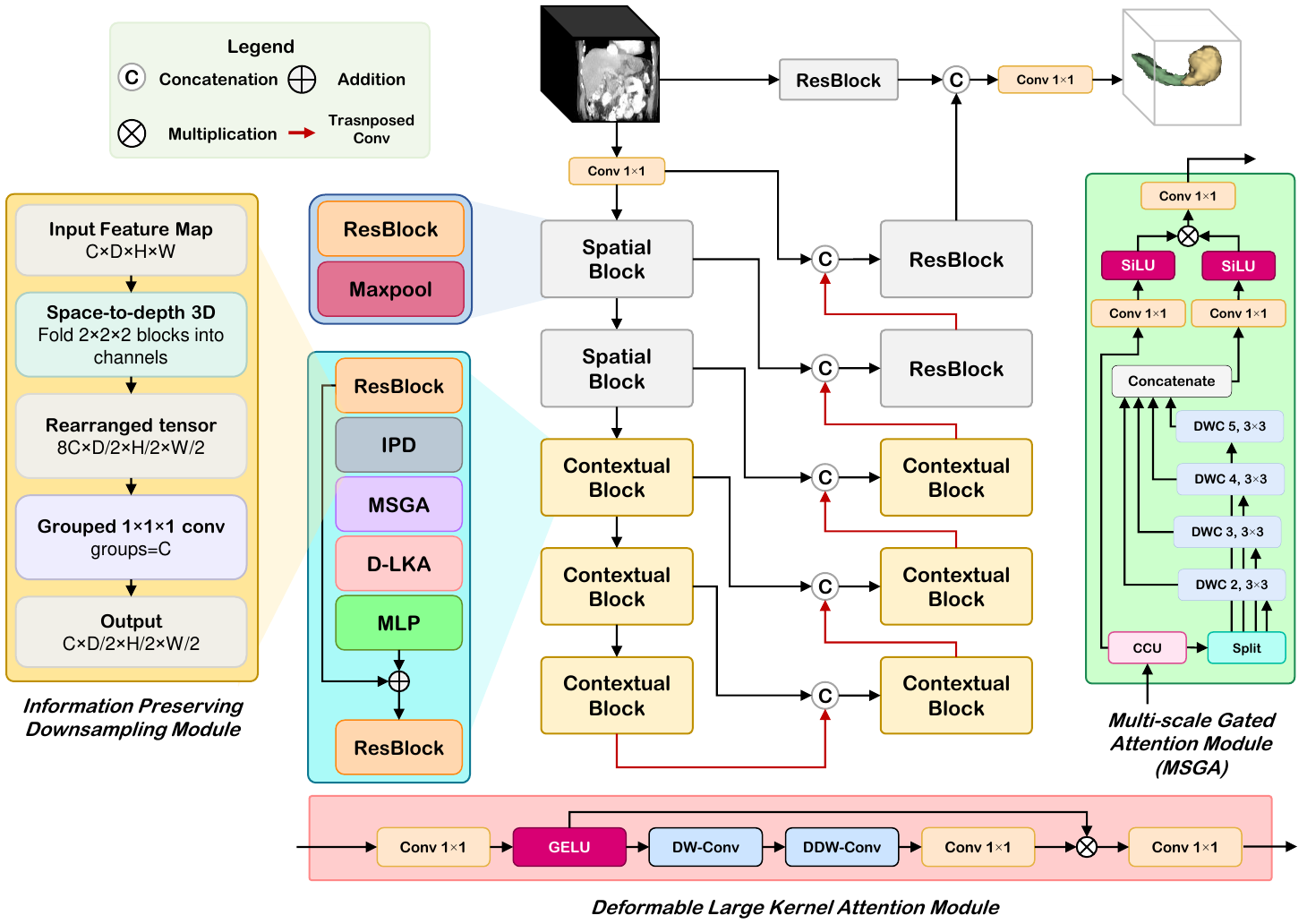}
  \caption{Overview of TRIUNE-Net.}
  \label{fig:arch}
\end{figure}

As shown in Fig.~\ref{fig:arch}, TRIUNE-Net is a U-shaped network. A
$1{\times}1{\times}1$ stem lifts the input volume
$x\in\mathbb{R}^{C_{\text{in}}\times D\times H\times W}$ into feature space; two
lightweight convolutional stages then reduce resolution, three semantic stages
perform the heavy lifting at coarser scales, and a mirror-image decoder restores resolution through transposed convolutions, concatenating encoder features at each stage via skip connections and ending in a 1×1×1 classifier. A residual full-resolution skip is concatenated before the classifier so that fine boundary detail bypasses the bottleneck entirely. Crucially, every downsampling step inside the semantic pathway is \emph{information-preserving} (Sec.~\ref{sec:down}) rather than max pooling, so a millimeter-scale lesion is never discarded on its way to the layers that can finally recognise it.

Each semantic stage is built from our \emph{Contextual Block}, a serial cascade of
three operators that confront the three challenges in turn. A Multi-Scale Gated
Aggregator (MSGA, Sec.~\ref{sec:msga}) reconciles the extreme \emph{scale}
disparity between organ and lesion; deformable large-kernel attention (D-LKA,
Sec.~\ref{sec:dlka}) adapts to irregular tumor \emph{form}; and a depthwise
convolutional MLP mixes the enriched features. We compose the operators in series
rather than in parallel, so that each refines an already context-aware
representation. The following subsections detail these three components and the
information-preserving downsampling that binds them across scales.

\subsection{Information-Preserving Downsampling}
\label{sec:down}

Each semantic stage begins by halving resolution. The usual choice, max pooling,
reduces every $2{\times}2{\times}2$ neighborhood to its maximum, keeping $1$ of $8$
voxels and discarding the other $7$; when a tumor spans only a handful of voxels,
this can erase the very signal we need. We instead use a simple, nearly
parameter-free alternative that throws nothing away. A \emph{space-to-depth}
operation~\cite{shi2016real} folds each $2{\times}2{\times}2$ block into the
channel dimension,
\begin{equation}
  \mathrm{S2D}:\ \mathbb{R}^{C\times D\times H\times W}\!\rightarrow\!
  \mathbb{R}^{8C\times \frac{D}{2}\times \frac{H}{2}\times \frac{W}{2}},
\end{equation}
after which a \emph{grouped} $1{\times}1{\times}1$ convolution (one group per output
channel) compresses $8C\!\rightarrow\!C$. Its weights are initialized to $1/8$, so
the layer starts as plain average pooling and then learns which sub-voxels to
emphasize. This halves each spatial dimension for only a handful of parameters per
stage, yet \emph{no voxel is discarded}: spatial information is reorganized rather
than dropped.

\subsection{Multi-Scale Gated Aggregation}
\label{sec:msga}

To perceive organ-scale and lesion-scale structure within one block, the MSGA
distributes the receptive field across channel groups. Given a normalized feature
map $x\in\mathbb{R}^{C\times D\times H\times W}$, a channel-calibration unit (CCU)
first reweights channels using global statistics: pooling the volume by maximum,
mean, and standard deviation yields a compact descriptor
$u\in\mathbb{R}^{C\times 3}$ that, through grouped $1$D convolutions, produces
per-channel gates $g_{c}=\sigma(\cdot)$, calibrating the response in the spirit of
squeeze-and-excitation~\cite{hu2018squeeze}:
\begin{equation}
  \tilde{x} = \mathrm{CCU}(x) = x \odot \sigma\!\big(W_2\,\rho(W_1\,
  [\,\mathrm{max}(x),\,\mathrm{mean}(x),\,\mathrm{std}(x)\,])\big).
\end{equation}
The calibrated features are split into $G$ groups along the channel dimension, one
per dilation rate, and each group $\tilde{x}^{(i)}$ is processed by a depthwise
$3{\times}3{\times}3$ convolution with dilation $d_i$, so that different groups
cover different receptive fields. The outputs are concatenated and fused through a
gated value projection:
\begin{equation}
  M = \big\Vert_{i=1}^{G}\, \mathrm{DWConv}_{3,\,d_i}\!\big(\tilde{x}^{(i)}\big),
  \qquad
  y = W_f\big(\,\mathrm{SiLU}(W_g\,\tilde{x}) \odot \mathrm{SiLU}(W_v M)\,\big) + x,
\end{equation}
where $\Vert$ denotes channel concatenation and all projections are
$1{\times}1{\times}1$ convolutions. Crucially, the dilation set is
\emph{stage-adaptive}, matched to the feature-map size of each semantic stage so
that the sampled receptive fields stay commensurate with the spatial extent that
remains. At the finest semantic stage ($1/8$ resolution) we use dilations
$\{2,3,4,5\}$ to span a wide range of scales; at $1/16$ resolution we use
$\{1,2\}$; and at the coarsest $1/32$ stage, where even a small kernel already
covers most of the volume, a single dilation $\{1\}$ suffices.
\subsection{Deformable Large-Kernel Attention}
\label{sec:dlka}

To model both context and irregular shape, we adopt deformable large-kernel
attention (D-LKA)~\cite{azad2024beyond}, a gated formulation in the spirit of
large-kernel attention~\cite{guo2023visual}. An input projection and GELU activation
produce a query map, which is modulated elementwise by a spatial gating branch
$A(\cdot)$:
\begin{equation}
  \mathrm{D\text{-}LKA}(x) = \phi(W_p\,x) \odot A\big(\phi(W_p\,x)\big),
  \qquad \phi=\mathrm{GELU}.
\end{equation}
The gating branch first applies a depthwise convolution for local context, then a
true 3D \emph{deformable} convolution~\cite{ying2020deformable} whose learned offsets let
the sampling grid bend to the tumor's contour, and finally a $1{\times}1{\times}1$
channel mixer:
\begin{equation}
  A(z) = W_m \, \mathrm{DeformConv}_{3}\big(\mathrm{DWConv}_{k}(z)\big).
\end{equation}
The kernel size $k$ is set proportionally to the channel width to balance local context and computational cost. The deformable convolution is placed directly after the depthwise context path, so its learned offsets operate on an already context-enriched representation and can bend the sampling grid toward irregular tumor boundaries.

\subsection{Convolutional MLP}
\label{sec:mlp}

Each Contextual Block closes with a depthwise-convolutional MLP that consolidates
the features produced by MSGA and D-LKA. After pre-normalization, an inverted
bottleneck expands the channel width fourfold with a $1{\times}1{\times}1$
convolution, applies a depthwise $3{\times}3{\times}3$ convolution that reinstates
local spatial structure, and projects back, all under a residual connection:
\begin{equation}
  \mathrm{MLP}(x) = x + W_2\,\psi\!\big(\mathrm{DWConv}_{3}\,\psi(W_1\,\mathrm{LN}(x))\big),
  \qquad \psi=\mathrm{GELU},
\end{equation}
where $W_1$ and $W_2$ are $1{\times}1{\times}1$ convolutions. The depthwise stage
gives the otherwise pointwise MLP a local receptive field, refining the
representation at negligible cost before the next block.

\subsection{Datasets and Data Splits}
\label{sec:datasets}

We evaluate on two datasets: the MSD Pancreas task (Task07) of the Medical Segmentation Decathlon~\cite{antonelli2022medical}, comprising $281$ portal-venous-phase abdominal CT volumes with voxel-level pancreas and tumor annotations ($224$ for training, $57$ for testing), and a synthetic dataset of $500$ CT volumes with corresponding segmentation masks generated using MAISI~\cite{guo2025maisi}, a 3D latent-diffusion model, which we name \textbf{NVD Pancreas} and split into $400/100$ for training and testing. For both datasets, we clip intensities to a soft-tissue window of $[-120,240]$\,HU and center-crop around the pancreas with a $25$-voxel margin.

\subsection{Implementation and Compute Resources}
\label{sec:compute}

TRIUNE-Net is a 3D U-shaped architecture trained from scratch with no external pre-training. We optimize with SGD (initial learning rate $1\!\times\!10^{-2}$, Nesterov momentum $0.99$, weight decay $3\!\times\!10^{-5}$) under a polynomial schedule for $15,000$ iterations, using $96{\times}96{\times}96$ patches, batch size $8$, standard augmentations (random rotations, scaling, and intensity shifts), and a combined Dice and cross-entropy loss. Training runs on four NVIDIA H100 GPUs (80GB each), completing in 45 minutes. For inference, we use a sliding-window protocol ($96{\times}96{\times}96$ patches, 50\% overlap, stride $48{\times}48{\times}48$) on a single H100 GPU. With only 5.86M parameters and 21.7 GFLOPs, a typical MSD Pancreas volume processes in approximately 1.7 seconds. We report Dice for pancreas and tumor, and additionally report sensitivity, precision, and F1 for tumor to capture the clinical costs of false negatives and false positives.
\section{Results}
\begin{figure}[t]
\centering
\begin{minipage}{0.49\linewidth}
    \centering
    \includegraphics[width=\linewidth]{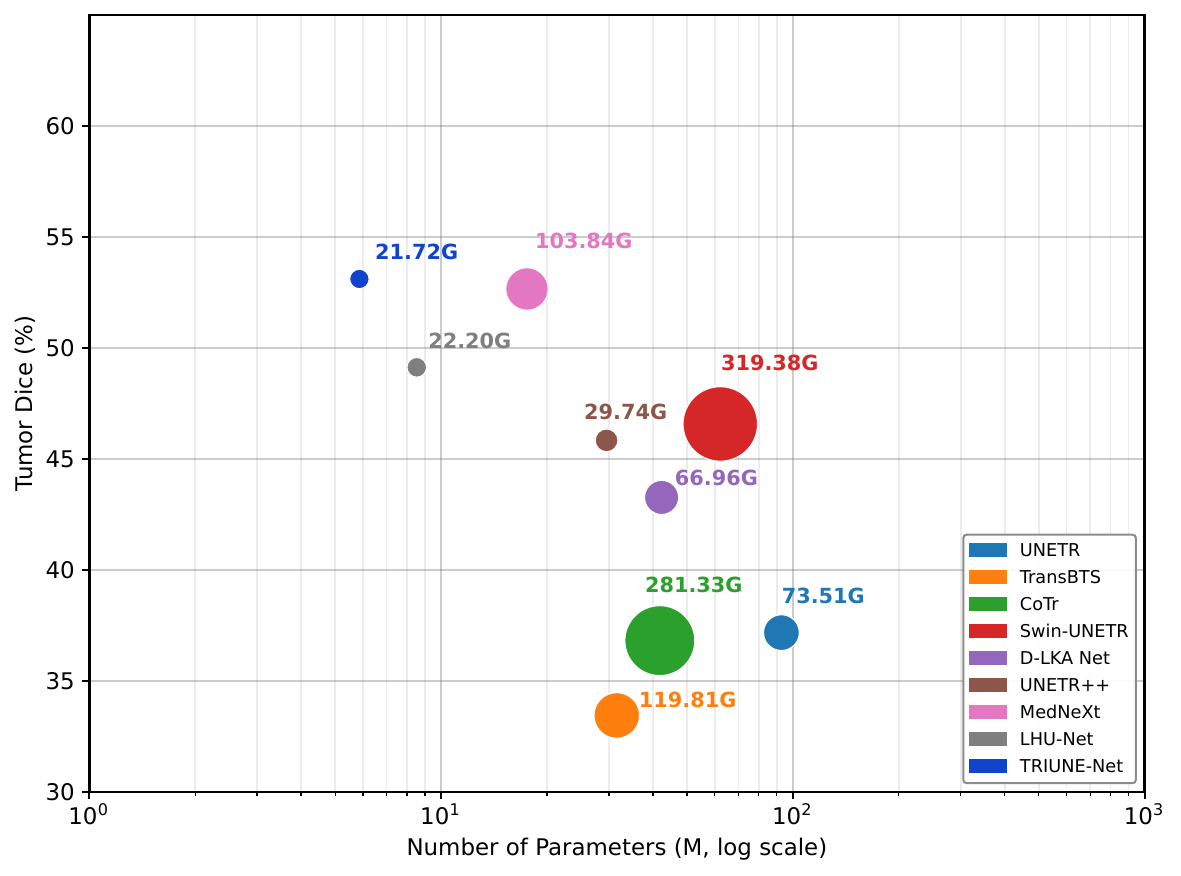}
    \caption{Tumor Dice vs computational cost.}
    \label{fig:efficiency}
\end{minipage}
\hfill
\begin{minipage}{0.49\linewidth}
    \centering
    \includegraphics[width=\linewidth]{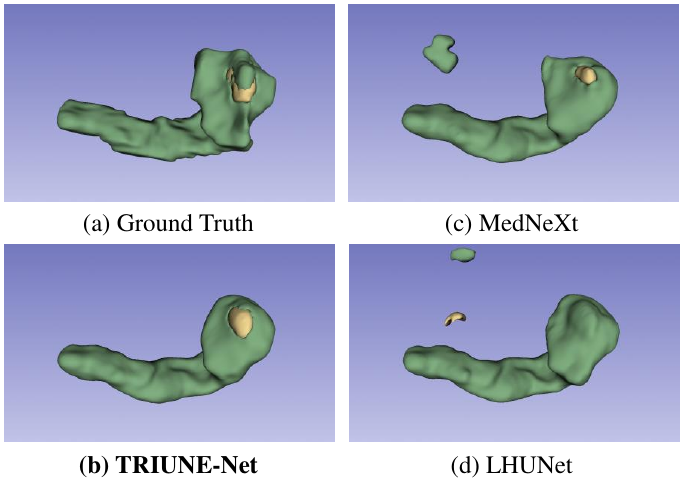}
    \caption{Qualitative comparison on MSD Pancreas.}
    \label{fig:qual}
\end{minipage}
\end{figure}

\begin{figure}[t]
\centering
\includegraphics[width=\linewidth]{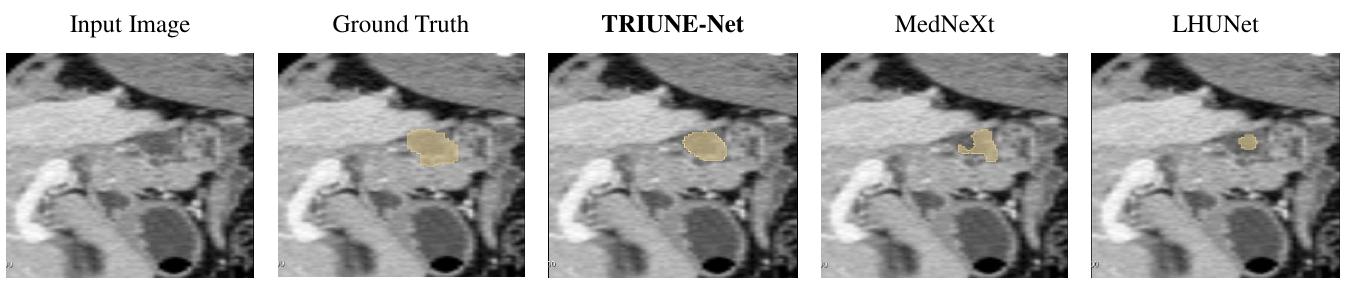}
\caption{2D axial slice comparison.}
\label{fig:qual2d}
\end{figure}

Table~\ref{tab:main} compares TRIUNE-Net with eight 3D segmentation models on 
MSD Pancreas. Two trends emerge. First, parameter count poorly predicts accuracy: 
heavy transformers trail compact designs, especially on tumors. Second, gains 
concentrate where the task is hardest. While top baselines cluster near 80\% 
pancreas Dice, tumor metrics separate them: TRIUNE-Net reaches 53.11\% tumor 
Dice, ahead of MedNeXt (52.66\%), with gains of +6.0 F1 over MedNeXt, +6.6 
sensitivity over LHU-Net, and +3.4 precision over MedNeXt. It does so with 
the smallest footprint (Fig.~\ref{fig:efficiency}). Qualitative results 
(Fig.~\ref{fig:qual}, Fig.~\ref{fig:qual2d}) show TRIUNE-Net closely follows 
ground truth boundaries, while MedNeXt produces fragmented predictions and 
LHU-Net severely undersegments tumors.

\begin{table}[t]
\centering
\caption{MSD Pancreas performance. Best in \textbf{\textcolor{blue}{blue}}, second-best in \textbf{\textcolor{red}{red}}.}
\label{tab:main}
\begin{tabular}{lccccccccc}
\toprule
\multirow{2}{*}{Method} & \multirow{2}{*}{Params (M)} & \multirow{2}{*}{FLOPs (G)} & \multicolumn{2}{c}{Pancreas} & \multicolumn{4}{c}{Tumor} \\
\cmidrule(lr){4-5} \cmidrule(lr){6-9}
 & & & DSC & F1 & DSC & F1 & Sens & Prec \\
\midrule
UNETR~\cite{hatamizadeh2022unetr} & 92.78 & 73.51 & 69.74 & 69.53 & 37.18 & 39.75 & 35.28 & 45.53 \\
TransBTS~\cite{wang2021transbts} & 31.58 & 119.81 & 73.21 & 73.08 & 33.45 & 35.95 & 31.87 & 41.23 \\
CoTr~\cite{xie2021cotr} & 41.86 & 281.33 & 75.44 & 75.31 & 36.82 & 39.19 & 34.91 & 44.68 \\
Swin-UNETR~\cite{hatamizadeh2021swin} & 62.19 & 319.38 & 79.13 & 79.23 & 46.58 & 48.04 & 43.78 & 53.22 \\
D-LKA Net~\cite{azad2024beyond} & 42.35 & 66.96 & 78.63 & 78.74 & 43.27 & 46.23 & 41.53 & 52.14 \\
UNETR++~\cite{shaker2024unetr++} & 29.54 & 29.74 & 79.38 & 79.51 & 45.84 & 49.32 & 43.12 & 55.67 \\
MedNeXt~\cite{roy2023mednext} & 17.55 & 103.84 & \textcolor{red}{\textbf{80.74}} & 80.86 & \textcolor{red}{\textbf{52.66}} & \textcolor{red}{\textbf{58.94}} & 50.11 & \textcolor{red}{\textbf{71.55}} \\
LHU-Net~\cite{sadegheih2024lhu} & \textcolor{red}{\textbf{8.53}} & \textcolor{red}{\textbf{22.20}} & 80.59 & \textcolor{red}{\textbf{80.98}} & 49.13 & 54.55 & \textcolor{red}{\textbf{50.67}} & 59.07 \\
\midrule
\textbf{TRIUNE-Net} & \textcolor{blue}{\textbf{5.86}} & \textcolor{blue}{\textbf{21.72}} & \textcolor{blue}{\textbf{80.84}} & \textcolor{blue}{\textbf{81.45}} & \textcolor{blue}{\textbf{53.11}} & \textcolor{blue}{\textbf{64.95}} & \textcolor{blue}{\textbf{57.30}} & \textcolor{blue}{\textbf{74.94}} \\
\bottomrule
\end{tabular}
\end{table}

On the synthetic NVD Pancreas dataset (Table~\ref{tab:nvd}), TRIUNE-Net again 
leads across all metrics: 86.54\% pancreas Dice and 59.21\% tumor Dice, with 
gains of +4.53 F1 and +7.13 sensitivity over MedNeXt, confirming robust 
generalization of the multi-scale design.

\begin{table}[t]
\centering
\caption{NVD Pancreas performance. Best in \textbf{\textcolor{blue}{blue}}, second-best in \textbf{\textcolor{red}{red}}.}
\label{tab:nvd}
\begin{tabular}{lccccccccc}
\toprule
\multirow{2}{*}{Method} & \multirow{2}{*}{Params (M)} & \multirow{2}{*}{FLOPs (G)} & \multicolumn{2}{c}{Pancreas} & \multicolumn{4}{c}{Tumor} \\
\cmidrule(lr){4-5} \cmidrule(lr){6-9}
 & & & DSC & F1 & DSC & F1 & Sens & Prec \\
\midrule
UNETR~\cite{hatamizadeh2022unetr} & 92.78 & 73.51 & 77.12 & 79.21 & 43.18 & 46.11 & 41.22 & 52.32 \\
TransBTS~\cite{wang2021transbts} & 31.58 & 119.81 & 78.45 & 78.56 & 39.65 & 42.70 & 37.98 & 48.76 \\
CoTr~\cite{xie2021cotr} & 41.86 & 281.33 & 80.82 & 80.91 & 43.12 & 45.93 & 41.23 & 51.84 \\
Swin-UNETR~\cite{hatamizadeh2021swin} & 62.19 & 319.38 & 84.67 & 84.78 & 52.38 & 54.22 & 49.62 & 59.78 \\
D-LKA Net~\cite{azad2024beyond} & 42.35 & 66.96 & 84.21 & 84.34 & 49.27 & 52.45 & 47.45 & 58.65 \\
UNETR++~\cite{shaker2024unetr++} & 29.54 & 29.74 & 84.96 & 85.12 & 51.84 & 54.79 & 49.12 & 61.95 \\
MedNeXt~\cite{roy2023mednext} & 17.55 & 103.84 & \textcolor{red}{\textbf{86.43}} & \textcolor{red}{\textbf{86.70}} & \textcolor{red}{\textbf{58.78}} & \textcolor{red}{\textbf{64.56}} & 56.32 & \textcolor{red}{\textbf{75.65}} \\
LHU-Net~\cite{sadegheih2024lhu} & \textcolor{red}{\textbf{8.53}} & \textcolor{red}{\textbf{22.20}} & 86.28 & 86.67 & 56.24 & 61.11 & \textcolor{red}{\textbf{56.82}} & 66.12 \\
\midrule
\textbf{TRIUNE-Net} & \textcolor{blue}{\textbf{5.86}} & \textcolor{blue}{\textbf{21.72}} & \textcolor{blue}{\textbf{86.54}} & \textcolor{blue}{\textbf{87.18}} & \textcolor{blue}{\textbf{59.21}} & \textcolor{blue}{\textbf{71.09}} & \textcolor{blue}{\textbf{63.45}} & \textcolor{blue}{\textbf{80.84}} \\
\bottomrule
\end{tabular}
\end{table}

\subsection{Ablation Study}


\begin{table}[t]
\centering
\begin{minipage}{0.48\linewidth}
\centering
\caption{Ablation on MSD Pancreas.}
\label{tab:abl}
\small
\begin{tabular}{ccc cc}
\toprule
\multirow{2}{*}{\centering MSGA} & \multirow{2}{*}{\centering D-LKA} & \multirow{2}{*}{\centering IPD} & \multicolumn{2}{c}{DSC (\%)} \\
\cmidrule(lr){4-5}
 & & & Pancreas & Tumor \\
\midrule
\ding{55} & \ding{55} & \ding{55} & 80.12 & 45.03 \\
\ding{51} & \ding{55} & \ding{55} & 80.35 & 49.70 \\
\ding{51} & \ding{51} & \ding{55} & 80.59 & 51.47 \\
\ding{51} & \ding{51} & \ding{51} & \textbf{80.84} & \textbf{53.11} \\
\bottomrule
\end{tabular}
\end{minipage}
\hfill
\begin{minipage}{0.48\linewidth}
\centering
\caption{Small tumor ($<$0.05).}
\label{tab:small_tumor_perf}
\small
\begin{tabular}{lcccc}
\toprule
Method & DSC & Sens & Prec & F1 \\
\midrule
MedNeXt & 35.9 & 47.5 & 28.5 & 35.6 \\
LHU-Net & 31.3 & 41.4 & 10.6 & 16.8 \\
SwinUNETR & 31.0 & 33.9 & 5.2 & 9.0 \\
UNETR & 14.1 & 23.9 & 5.8 & 9.3 \\
\textbf{TRIUNE} & \textbf{37.5} & \textbf{52.4} & \textbf{29.1} & \textbf{37.4} \\
\bottomrule
\end{tabular}
\end{minipage}
\end{table}

We conduct two ablations on the MSD Pancreas dataset: (1) component contributions (Table~\ref{tab:abl}), and (2) performance on small tumors (Table~\ref{tab:small_tumor_perf}) with relative size $<0.05$, computed as $\text{Tumor}/(\text{Tumor}+\text{Pancreas})$, capturing early-stage lesions most critical for patient outcomes.
MSGA provides the largest gain (+4.67\% tumor DSC), confirming multi-scale 
context aggregation is critical for handling extreme scale disparity. 
D-LKA and IPD contribute +1.77\% and +1.64\% respectively, addressing 
complementary challenges of irregular morphology and information preservation. 
Monotonic improvements across configurations demonstrate all three components 
are necessary.
On small tumors, TRIUNE-Net achieves the highest scores across all metrics, 
with the largest gap in sensitivity (+4.9 over MedNeXt) while maintaining 
higher precision (29.1\% vs 28.5\%). This simultaneous improvement indicates 
TRIUNE-Net detects more early-stage lesions without increasing false positives, 
reducing unnecessary follow-ups while improving diagnostic yield.

\section{Discussion}
We presented TRIUNE-Net, a lightweight architecture for pancreatic tumor 
segmentation. Multi-scale context aggregation with stage-adaptive dilated 
convolutions handles broad scale variability, serial linear-deformable 
attention adapts to irregular non-convex morphologies, and an information-
preserving downsampling module retains spatial information conventional max 
pooling would discard. On MSD Pancreas, TRIUNE-Net achieves 80.84\% pancreas 
and 53.11\% tumor Dice, outperforming all baselines on sensitivity and 
precision, making it practical for opportunistic screening in resource-
constrained settings. Future work includes validating the approach on 
additional small-lesion tasks and larger datasets with both cancerous and 
non-cancerous cases, better reflecting real-world clinical deployment.

\section{Impact in Resource-Constrained Settings}
TRIUNE-Net is practical for resource-limited deployment. At 5.86M parameters 
and 21.7 GFLOPs, it is the smallest model evaluated, processing a full volume 
in ~1.7 seconds on a single GPU without dedicated AI servers or cloud 
infrastructure. This efficiency does not trade off against safety: TRIUNE-Net 
improves sensitivity and precision simultaneously, including on small, 
early-stage tumors (+4.9 sensitivity over MedNeXt with no precision drop), 
catching more curable tumors while limiting unnecessary follow-up. Training 
from scratch, without external pre-trained weights, further supports data 
sovereignty, letting institutions adapt the model locally without sharing 
patient data.

\bibliographystyle{unsrt}

\bibliography{refs}

\end{document}